\documentclass[acmsmall,authorversion,nonacm]{acmart}
\AtBeginDocument{%
  }
\authorsaddresses{}

\usepackage{CJKutf8}

\usepackage{amssymb}
\usepackage{xcolor}
\newboolean{showcomments}
\setboolean{showcomments}{true}
\ifthenelse{\boolean{showcomments}}
{ \newcommand{\mynote}[3]{
		\fbox{\bfseries\sffamily\scriptsize#1}
		{\small$\blacktriangleright$\textsf{\emph{\color{#3}{#2}}}$\blacktriangleleft$}}}
{ \newcommand{\mynote}[3]{}}
\newcommand{\shrink}[1]{}
\definecolor{pink}{rgb}{1,0.2,0.7}
\definecolor{purple}{rgb}{0.7,0,0.9}

\begin{document}

\begin{CJK}{UTF8}{bsmi}
\title{The Reliability Issue in ReRam-based CIM Architecture for SNN: A Survey}


\author{Wei-Ting, Chen}
\affiliation{%
  \institution{National Taiwan University}
  \city{Taipei}
  \country{Taiwan}}
\email{b08508024@ntu.edu.tw}


\begin{abstract}
  The increasing complexity and energy demands of deep learning models have highlighted the limitations of traditional computing architectures, especially for edge devices with constrained resources. Spiking Neural Networks (SNNs) offer a promising alternative by mimicking biological neural networks, enabling energy-efficient computation through event-driven processing and temporal encoding. Concurrently, emerging hardware technologies like Resistive Random Access Memory (ReRAM) and Compute-in-Memory (CIM) architectures aim to overcome the Von Neumann bottleneck by integrating storage and computation. This survey explores the intersection of SNNs and ReRAM-based CIM architectures, focusing on the reliability challenges that arise from device-level variations and operational errors. We review the fundamental principles of SNNs and ReRAM crossbar arrays, discuss the inherent reliability issues in both technologies, and summarize existing solutions to mitigate these challenges.
\end{abstract}

\maketitle

\section{Introduction}
The concept of artificial neural networks (ANNs) has been present since the mid-20th century, with early foundational work by McCulloch and Pitts in 1943. However, despite promising theoretical developments, the progress of neural networks remained limited for several decades due to substantial hardware constraints. Early models faced significant challenges, such as insufficient computational power, limited memory, and a lack of efficient training algorithms. These barriers largely restricted the practical application and scalability of neural networks, leading to an "AI winter" during which research interest waned.

It was not until the early 21st century, with the rapid advancement of hardware technologies such as Graphics Processing Units (GPUs) and the development of distributed computing, that neural networks began to realize their potential. The surge in computational power, combined with large-scale datasets and novel algorithms, enabled the training of much deeper architectures, paving the way for the modern era of deep learning. Techniques such as backpropagation, which had been conceptually understood for decades, finally became computationally feasible, allowing for effective training of multi-layer networks.

Over the past two decades, the field of deep learning has seen continuous evolution, beginning with the introduction of the deep Convolutional Neural Network, AlexNet, in 2012. Since then, numerous algorithms and architectures for deep learning models such as VGGNet\cite{DBLP:journals/corr/SimonyanZ14a}, ResNet\cite{7780459}, BERT\cite{Devlin2019BERTPO}, and GPT-3\cite{10.5555/3495724.3495883} have been proposed. These deeper model architectures have improved models’ capabilities.

However, these deep learning (DL) model architectures demand an increased number of model parameters, higher complexity, and longer training times, even when utilizing the most powerful and advanced hardware, leading to significantly higher energy consumption. This means that unless hardware continues to advance at its current pace, as described by Moore's Law, DL models will eventually reach computational limits due to hardware constraints. Additionally, certain devices that require low energy consumption, such as edge devices, face challenges in adopting these advanced DL models. Consequently, in the past decade, research has shifted towards new hardware architectures and novel neuron models aimed at simultaneously enhancing task accuracy and efficiency while reducing model parameters.

Most conventional processors are based on the Von Neumann architecture, in which the processor and memory are separate units. As a result, the processor needs to access instructions and data from memory, leading to the well-known Von Neumann bottleneck\cite{vn}. This bottleneck arises because the data transfer rate between the processor and memory is significantly lower compared to the computational speed of the processor. Moreover, as processors have become faster, the disparity between processor speed and memory access speed has increased, exacerbating the bottleneck issue.

To address this issue, the concept of Compute-in-Memory (CIM) has been proposed. CIM refers to a new system architecture that performs computation directly within memory, eliminating the need to transfer data to separate processing units and thereby saving substantial data transfer time. This architecture helps overcome the Von Neumann bottleneck. Non-volatile memory (NVM) technologies, such as Resistive Random Access Memory (ReRAM) and Phase-Change Memory (PCM), can be utilized in CIM architectures. These emerging hardware technologies have physical properties that can reduce the time and energy required for operations in neuron models. For example, the physical characteristics of ReRAM can reduce the time complexity of vector-matrix multiplication (VMM) operations from O($n^2$) to O(1).

In addition to novel hardware architectures, researchers are also exploring new algorithms and neuron models, such as the Spiking Neural Network (SNN). The core concept behind SNNs is to mimic the physiological properties of neuronal cells, where communication between neurons is based on neuronal "spikes" or firing events in the brain. This feature contributes to SNNs' greater energy efficiency compared to traditional deep learning models. Furthermore, SNNs possess characteristics such as implicit recurrence, event-driven computation, inherent sparsity, and temporal encoding, all derived from their all-or-nothing spike operation mode.

To date, numerous studies have integrated SNNs with emerging hardware technologies \cite{10.1145/2742060.2743756} \cite{10.1145/3061639.3062311} \cite{demirag2021onlinetrainingspikingrecurrent}, revealing that the results tend to exhibit slightly lower accuracy than traditional deep learning models but can significantly reduce energy consumption and enhance model efficiency. These characteristics make the combination of SNNs and new hardware architectures particularly suitable for use in edge devices. However, some of the existing research has been conducted in simulated environments, and the integration of SNNs with new hardware in real-world settings may introduce errors due to their physical characteristics, leading to reliability issues. Therefore, this survey paper aims to explore the operating principles of ReRAM and SNNs, examine the reliability challenges they face, and review previous research to summarize potential solutions for addressing these reliability issues.

\section{Spiking Neural Network}
The concept of Spiking Neural Networks (SNNs) is derived from the operational patterns of neurons within biological organisms, often referred to as biological neuron models or spiking neuron models. The functionality of neurons in living organisms is inherently complex, with each type of neuron exhibiting specific operational characteristics yet sharing certain commonalities. Simplifying neural operation for easier understanding, it can be divided into three main components: the pre-synaptic neuron, the synapse, and the post-synaptic neuron. The pre-synaptic neuron connects to another neuron through the synapse (refer to Figure 1). The operational mode of a single neuron involves receiving spikes transmitted from preceding neurons. The timing and frequency of these spikes cause changes in the neuron's membrane potential. When this membrane potential exceeds a certain threshold, the neuron will emit a spike (see Figure 2).

\begin{figure}[h]
    \centering
    \includegraphics[width=1\textwidth]{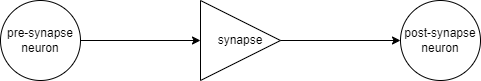}
    \caption{A synapse connects the pre-synaptic neuron to the post-synaptic neuron}
    \label{fig:ppsy}
\end{figure}
\begin{figure}[h]
    \centering
    \includegraphics[width=1\textwidth]{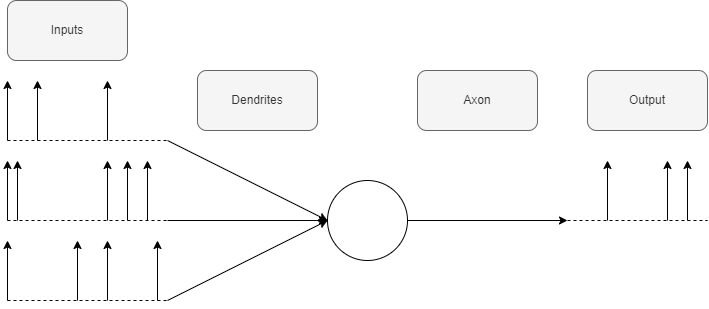}
    \caption{The input signals are received by dendrites, processed in the cell body, when membrane potential exceeds a
certain threshold, the neuron will emit a spike.}
    \label{fig:sn}
\end{figure}
Like other cells, a neuron's membrane is composed of a bilayer of phospholipids, which includes selective channels allowing ions to pass through. Physically, the bilayer of phospholipids acts as a capacitor, while the ion channels form resistances. The differing concentrations of ions inside and outside the cell create a voltage difference across the membrane. With these properties, it is possible to represent a neuron in the form of an electrical circuit. Figure 3 illustrates the simplest form of such a circuit: the Integrate-and-Fire (I\&F) circuit.

In 1952, Hodgkin and Huxley published a paper in which they studied the operational dynamics of the giant squid's axon nerve cells\cite{Al1952AQD}. Through physiological experiments, they were able to simplify the operational mode of these nerve cells into an electrical circuit form, depicted in Figure 4. In their model, the cell's phospholipid bilayer is represented by a capacitance $C_{m}$. The sodium and potassium ion channels are collectively represented by a voltage source and a conductance (indicated in the diagram as $E_{n}$ and $g_{n(t, V))}$, while the chloride ion channels are also represented by a voltage source and a conductance (indicated as $E_{L}$ and $g_{L(t, V))}$. The ion pumps and exchangers are represented by current sources $I_{p}$, and the membrane potential is represented by $V_{m}$. This model was a groundbreaking contribution to neurophysiology, providing a quantitative description of how nerve cells generate electrical signals.

\begin{figure}[htb]
\centering
\begin{minipage}{0.48\linewidth}
\centering
\includegraphics[width=1\textwidth]{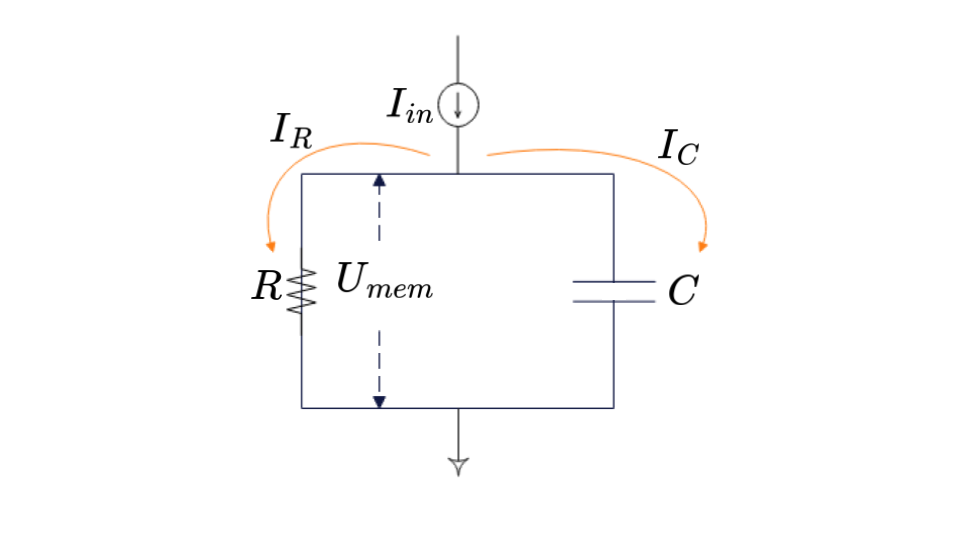}
\caption{Equivalent circuit model of a leaky integrate-and-fire (LIF) neuron.}
\label{fig:IF_circuit}
\end{minipage}\hfill
\begin{minipage}{0.48\linewidth}
\centering
\includegraphics[width=1\textwidth]{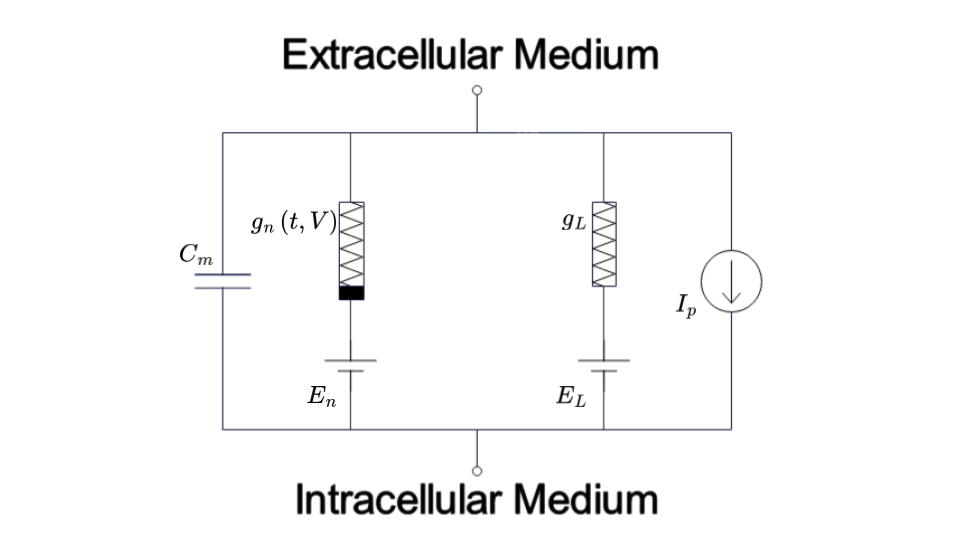}
\caption{A typical Hodgkin–Huxley model}
\label{fig:squid_cell}
\end{minipage}
\end{figure}

Because the Hodgkin-Huxley model directly mimics the behavior of biological neurons and considers the interactions among multiple ion channels, it is highly complex. This complexity makes it less suitable for direct application in neural networks. Currently, the most commonly used neuron models in spiking neuron networks are the Integrate-and-Fire (I\&F), the Leaky-Integrate-and-Fire (LIF) \cite{10.1145/3061639.3062311} \cite{doi:10.1073/pnas.1604850113}, and the Spike Response Model (SRM). These models simplify the dynamics of biological neurons to a degree that makes them more practical for implementation in neural networks while still capturing essential aspects of neuronal activity.

\subsection{LIF Neuron Model}
Figure 3 illustrates a basic LIF model. Assuming that at time t an input current $I_{in(t)}$ is introduced into the circuit, we get $I_{in(t)}=I_{R}+I_{C}$. According to Ohm's Law, the voltage difference across the cell membrane $U_{mem}$ is directly proportional to the current flowing through the resistor:

\begin{center}
$I_{R(t)}=U_{mem/R}$
\end{center}
The capacitance C is related to its charge Q and the voltage difference across the cell membrane $U_{mem}$:

\begin{center}
$Q=C*U_{mem(t)}$
\end{center}
Differentiating Q with respect to time, we obtain:

\begin{center}
$dQ/dt=I_{C(t)}=C*dU_{mem}/dt -> R*C*dU_{mem(t)}/dt=-U_{mem(t)}+R*I_{in(t)}$
\end{center}
Here, R*C is the time $constant(\tau)$ of the circuit, leading to the final formula:

\begin{center}
$\tau*dU_{mem(t)}/dt=-U_{mem(t)}+R*I_{in(t)}$
\end{center}

From the circuit, it is evident that the LIF model is much simpler than the Hodgkin-Huxley (H-H) model. While the LIF model cannot fully represent the complex dynamics of biologically accurate neurons,  its ability to approximate some of these dynamics under a simplified framework makes it particularly well-suited for the architecture of SNNs. This balance between simplicity and biological realism allows the LIF model to efficiently simulate neuron behavior in a way that is computationally feasible for large-scale neural networks, making it a popular choice in the field of neuromorphic computing.

In the LIF model, when input is received, the neuron's membrane potential accumulates, and this integration is leaky, meaning the membrane potential will gradually decay over time. When the neuron reaches its threshold, it outputs a spike. Sometimes a refractory period occurs immediately after the neuron fires a spike, during which time, even if there are input stimuli, the membrane potential will not increase. This prevents excessive firing by a single neuron. After firing, the neuron's membrane potential resets to the resting state following the refractory period. This cycle allows the LIF model to effectively simulate neuronal firing and resting phases in a computationally efficient manner.

\subsection{Spike Coding} 

SNNs typically require both their input and output to be in binary spike form. This necessitates the use of methods to encode regular data into spike signals. Using a binary format for encoding data offers several advantages. First, representing data as "0" or "1" (and in some models, "-1" for suppression) simplified representation in hardware compared to high-precision values, which can save processing time, area, and energy. \cite{9218689} \cite{10.1007/978-3-030-58607-2_23} \cite{10.1007/978-3-031-30105-6_48} \cite{10322581} Moreover, binary-formatted data exhibits significant sparsity: neurons will often not output spikes, meaning the neuron outputs are mostly "0". Many zero outputs (forming a sparse vector) are easier to store and manage, and when the output is zero, operations typically do not need to be executed, meaning there is no need to read from memory. This characteristic makes neuromorphic hardware much more efficient. The most commonly used encoding methods in SNNs include rate encoding, temporal encoding, and delta modulation.

\subsubsection{Rate Coding}:

In rate coding, information is represented by the number of spikes emitted within a fixed period \cite{Adrian1926TheIP} \cite{Denve2016EfficientCA}. For instance, to convert the number five into spike form, rate coding might generate five spikes within one second. When applying rate coding to image data, such as converting an image from the MNIST dataset into spikes, a Bernoulli trial function can be used for the transformation. This method treats each input feature $X_{ij}$ as a probability event at any given time, ultimately returning a rate-coded value $R_{ij}$.

Using the Bernoulli trial function, we can model $R_{ij} = B(n, p)$, where n = 1 (number of trials), and $p = X_{ij}$ (probability of success or spiking). That is, the probability that a spike occurs is $P(R_{ij} = 1) = X_{ij}$. This means that in an image, a white pixel (with the highest intensity) has a 100\% probability of outputting a spike, a black pixel has a 0\% probability, and a gray pixel (medium intensity) has a 50\% probability of outputting a spike.

In rate coding, the number of time steps influences accuracy; hence, using rate coding requires consideration of the accuracy-latency trade-off. The longer the time window allowed for observing spikes, the higher the potential accuracy, but at the cost of increased inference latency. \cite{Sengupta2018GoingDI}

Moreover, rate encoding distributes information probabilistically over a fixed time interval, potentially overlooking some of the temporal aspects that spikes can signify. This might limit the ability to capture temporal dynamics where the timing of events carries significant information. Thus, while rate encoding is straightforward and can effectively translate amplitude information into spike rates, it might not be the best choice for all types of data or applications where the temporal sequence or the exact timing of spikes is essential.

\subsubsection{Temporal Coding}:

In temporal encoding, the timing of each spike is a critical feature, encoding information based on when the spike occurs. The inter-spike interval considers the time differences between consecutive spikes from the same neuron as significant \cite{Oswald2007IntervalCI}. The intervals can encode varying levels of information depending on their duration, allowing the system to represent a range of values dynamically. Rank order coding encodes information based on the order in which spikes occur, rather than their specific timings \cite{Gautrais1998RateCV} \cite{inproceedingstsg}. Neurons that fire earlier encode larger analog values, reducing the need for precise timekeeping by focusing on the relative timing of spikes. Time-to-first-spike coding considers the timing of the first spike as the most informative, suggesting that the initial response time of the neuron to a stimulus holds the most significant data \cite{Gollisch2008RapidNC} \cite{Johansson2004FirstSI}.

Temporal encoding offers the advantage of encoding information with fewer spikes, which can be more energy-efficient. However, this efficiency can come at the cost of increased memory and computational requirements, as precise spike timing needs to be recorded and analyzed.

The choice of temporal coding method often depends on the specific application and the nature of the data being processed. For instance, tasks requiring rapid response might benefit from time-to-first-spike encoding, while those involving complex, time-sensitive patterns might better utilize inter-spike interval or rank order coding to capture the dynamics within the signal.

\subsubsection{Delta Modulation}:

In human visual perception, photoreceptors encode differences in light intensity, reflecting an adaptive characteristic of our vision system \cite{articledelta}. Human vision systems process information primarily when changes occur in our visual environment. If the visual scene remains unchanged, the photoreceptors will not fire new spikes. This phenomenon suggests that our vision effectively operates on a principle similar to delta modulation, where the encoding of data is based on the differences between consecutive values.

Delta modulation encodes data by assessing the difference between successive data points; if the difference in successive data is both positive and exceeds a certain threshold, a spike is generated. 

Event cameras such as Samsung DVS \cite{7870263}, DVS128 \cite{4444573}, DAVIS240 \cite{6889103} can directly capture changes in the scene asynchronously and independently, thereby encoding the delta-modulated information. event cameras have high temporal resolution, high dynamic range, and low power consumption compared to standard cameras.

Using event cameras with delta modulation is highly efficient for dynamic environments, reducing the volume of data transmitted and processed, and lowering power consumption.

These encoding strategies are crucial for converting analog and digital inputs into the binary spike format required by SNNs, thus leveraging the inherent advantages of neuromorphic computing such as energy efficiency and reduced hardware demands. Ultimately, the trade-offs between complexity, latency, energy efficiency, and accuracy must be carefully considered to select the most appropriate encoding strategy for a given neuromorphic computing task or neural network implementation.

\subsection{Learning Rules} 

The unique operation mode of SNNs and their distinct neuron models mean that some learning rules applicable to ANNs cannot be directly transferred to SNNs. Moreover, there are learning rules specifically designed for the spike-based architecture of SNNs. This section introduces some of the most commonly used learning rules in SNNs.

\subsubsection{Unsupervised Learning}:

Unsupervised learning is an algorithm that can find patterns from unlabeled data. In SNNs, a common method of unsupervised learning is Spike Timing Dependent Plasticity (STDP) \cite{Caporale2008SpikeTP} \cite{articlestdp} \cite{lu2024deepunsupervisedlearningusing} \cite{10.1016/j.neunet.2017.12.005}. Originally a biological mechanism for adjusting the strength of connections between neurons, the principle of STDP is that if a pre-synaptic neuron fires before a post-synaptic neuron, their connection is strengthened. Conversely, if the post-synaptic neuron fires before the pre-synaptic neuron, their connection is weakened. Strengthening a connection is known as long-term potentiation (LTP), while weakening a connection is referred to as long-term depression (LTD).

The STDP used in SNNs is derived from this biological concept, with one of the most common STDP formulas being:

\[
    \delta w = \{\begin{array}{lr}
        Ae^{-(|t_{pre}-t_{post}|)/\tau}, & \text{if } t_{pre}-t_{post}\leq 0, A \textgreater 0\\
        Be^{-(|t_{pre}-t_{post}|)/\tau}, & \text{if } t_{pre}-t_{post}\textgreater 0, B\textless 0\\
        \end{array}
\]
\cite{DanUnknownTitle2006}

In this formula, w represents the synaptic weight. A and B are the learning rates, and $\tau$ is the time constant used to set the learning time window. The first formula indicates that when the pre-synaptic neuron's spike timing is ahead of the post-synaptic neuron's spike timing, it results in strengthened connections, representing LTP behavior. The second formula represents LTD behavior, where the post-synaptic neuron's spike timing ahead of the pre-synaptic neuron's leads to weakened connections.

STDP is frequently utilized in pattern recognition tasks. Early studies showcased the potential of SNNs for unsupervised pattern recognition using a two-layer architecture. This architecture typically consists of an input layer connected to an excitatory layer, which in turn is linked to an inhibitory layer\cite{4370930}. However, such two-layer SNNs are generally limited to simpler tasks, like handwritten digit recognition.Subsequently, convolutional SNNs were introduced, mirroring the architecture of deep neural network (DNN) convolutional networks. Unlike DNNs that use backpropagation for training, convolutional layers in these SNNs are trained using STDP. Although the performance of convolutional SNNs surpasses that of fully connected two-layer SNNs, they still do not match the performance of ANNs on benchmark datasets.

To date, proposed convolutional STDP SNNs have only incorporated a few convolutional layers, leaving the effective depth of such networks largely unexplored \cite{Deng2020RethinkingTP}. If the accuracy of deep convolutional STDP SNNs could eventually match that of ANNs, the future development could pivot towards a low-cost, local, unsupervised learning framework that dynamically adapts based on environmental changes. This possibility highlights the need for further exploration into how deep convolutional STDP SNNs can be effectively structured and scaled.

\subsubsection{Supervised Learning}:

Unlike unsupervised learning, supervised learning requires labeled data for training. It receives input and outputs predictions, which are then compared with labeled data to compute a loss. Finally, network parameters are updated based on the gradients of the loss function. However, in SNNs, the spiking neurons have discontinuous and non-differentiable characteristics, so we need spike-based backpropagation to address this issue.

Assuming an SNN neuron model where the input data is X, synaptic current is I, membrane potential is U, and spiking output is S, with the loss function as L, then the forward pass would be X→I→U→S→L. But the S formula in the SNN neuron model is: 

\[
    S[t] = \{\begin{array}{lr}
        1, & \text{if } U[t] \textgreater U_{thr} (U_{thr} \text{ is membrane potential threshold}) \\
        0, & \text{otherwise }\\
        \end{array}
\]

which means $S[t]=\theta (U[t] $-$ U_{thr})$ where $\theta$ is the Heaviside step function. If backpropagation is to be applied to the SNN neuron model, we need to consider the value of $\delta S/ \delta U$. However, $\delta S/ \delta U$ is Dirac Delta function, the derivative of the Heaviside step function. This Dirac delta function only has a value at the threshold point and is zero at all other points, implying that the gradient at every point is zero, and at the threshold point, it will saturate. Therefore, the neuron model cannot learn, leading to the dead neuron problem.

To solve the dead neuron problem, the method of surrogate gradients was proposed, whose core concept is to transform the Heaviside step function of the SNN neuron model into an approximate but continuous function during the backward pass, such as the arctangent function, sigmoid function, or the fast sigmoid function. \cite{8891809} \cite{Boht2011ErrorBackpropagationIN} \cite{Neftci2019SurrogateGL} \cite{10.1162/neco_a_01086}

However, because the training process for SNNs requires consideration of time steps, a single forward pass in an ANN corresponds to multiple forward passes in an SNN. Additionally, the backward pass in SNNs requires the integration of gradients across the total number of time steps, thereby significantly increasing both the computational complexity and memory requirements. These demands for computational resources and memory limit its practical application to smaller datasets, such as MNIST and CIFAR-10.

This limitation significantly impacts the scalability of SNNs for handling more complex or larger datasets typically encountered in real-world applications. Consequently, while SNNs offer potential advantages in terms of power efficiency and biological realism, their broader adoption in deep learning tasks is hampered by these computational and resource constraints. Addressing these limitations might involve optimizing the efficiency of spike-based computations or developing new algorithms that reduce the number of required time steps or simplify the integration process during the backward pass.

\subsubsection{ANN to SNN Conversion}:

SNNs employing the learning rules mentioned above have shown good accuracy on simpler datasets, but their performance on more complex or challenging datasets has not been ideal. The method of converting an ANN to an SNN can enable SNNs to achieve accuracies very close to those of ANNs\cite{10.1007/s11263-014-0788-3} \cite{7280696} \cite{10.3389/fnins.2017.00682} \cite{NEURIPS2021_afe43465}. This approach involves initially training an ANN model using the ReLU activation function with certain restrictions, such as omitting biases and batch normalization. A similar architecture is then prepared for the SNN, and the weights of its I\&F neurons are initialized with those from the trained ANN. The firing thresholds of the SNN's I\&F neurons are adjusted to align their average firing rates with the activations of the ReLU function. The resulting converted SNN can achieve high accuracy on more complex datasets, albeit at the cost of high inference latency.
The high inference latency in ANN-to-SNN conversion arises because the conversion process does not fully utilize the temporal information inherent in SNNs. In contrast, spike-based backpropagation utilizes techniques like backpropagation through time (BPTT), allowing it to incorporate temporal information during the backpropagation process and achieve lower inference latency than ANN-to-SNN conversion.

\section{ReRam and Crossbar Array}

As advanced AI models continue to grow in complexity with an increasing number of parameters, their power consumption has significantly increased. This poses a challenge for edge devices, which are constrained by their hardware capabilities and cannot support AI models with high power demands. Consequently, there has been a growing emphasis on developing AI models and hardware architectures with lower power consumption and higher efficiency for edge devices. Among the various NVM technologies, Resistive Random Access Memory (ReRAM) with crossbar arrays has become very popular and promising.

ReRAM devices store data based on the resistance levels within the device, allowing them to perform both computing and memorizing tasks simultaneously. This dual capability is highly beneficial for enhancing computational efficiency. The crossbar arrays can accelerate operations commonly used in ANNs, such as multiply-accumulate (MAC) operations. Within these arrays, operations are executed in an analog manner by setting different input voltages and adjusting resistance values to represent the neural network’s inputs and weights.

The structure of ReRAM consists of a Metal-Insulator-Metal (MIM) configuration \cite{6193402}. The insulating layer, known as the resistive switching layer, is typically made from materials like HfOx, TiO2, Al2O3, or their combinations. Altering the resistance of this layer is achieved by applying a voltage (set or reset voltage). When a set voltage is applied, it leads to the formation of oxygen vacancies, creating a conductive filament that allows electrons to hop on, placing the ReRAM in a Low Resistance State (LRS). Conversely, the reset process involves applying a reverse polarity voltage, often a negative voltage, which breaks the conductive filament by creating a gap between the oxygen particles and the metal layers. This disrupts the path for electrons, resulting in a High Resistance State (HRS).

A crossbar array is composed of word lines (WLs) and bit lines (BLs), arranged perpendicularly in a 3D grid-like structure, with non-volatile memory located at the intersections of these lines. There have already been many studies utilizing this structure \cite{10.1145/3061639.3062311} \cite{7551379}. An example of such a setup can be illustrated with a ReRAM-based crossbar accelerator (see Figure 5).

The operation of a ReRAM-based crossbar accelerator begins with programming the neural network’s parameters into the ReRAM’s conductance levels. Input data are then converted from digital to analog signals by a digital-to-analog converter (DAC) and fed into the WLs. The currents flowing through the BLs produce the results of the MAC operations. Finally, these currents are converted back into digital signals by an analog-to-digital converter (ADC).

Taking the example in Figure 5, the input data are first converted into voltages $V_{1}$ and $V_{2}$ by the DAC. These voltages are then applied to $WL_{1}$ and $WL_{2}$. According to Kirchhoff's Current Law (KCL), the currents $I_{1}$, $I_{2}$, $I_{3}$, and $I_{4}$ at the intersections are determined by the voltages $V_{1}$ and $V_{2}$ as well as the conductances $G_{1}$, $G_{2}$, $G_{3}$, and $G_{4}$ that have been programmed into the ReRAM. The total currents on the bit lines are then summed to compute the results of the MAC operations: $I_{5} = I_{1} + I_{2}$ and $I_{6} = I_{3} + I_{4}$

\begin{figure}[h]
    \centering
    \includegraphics[width=1\textwidth]{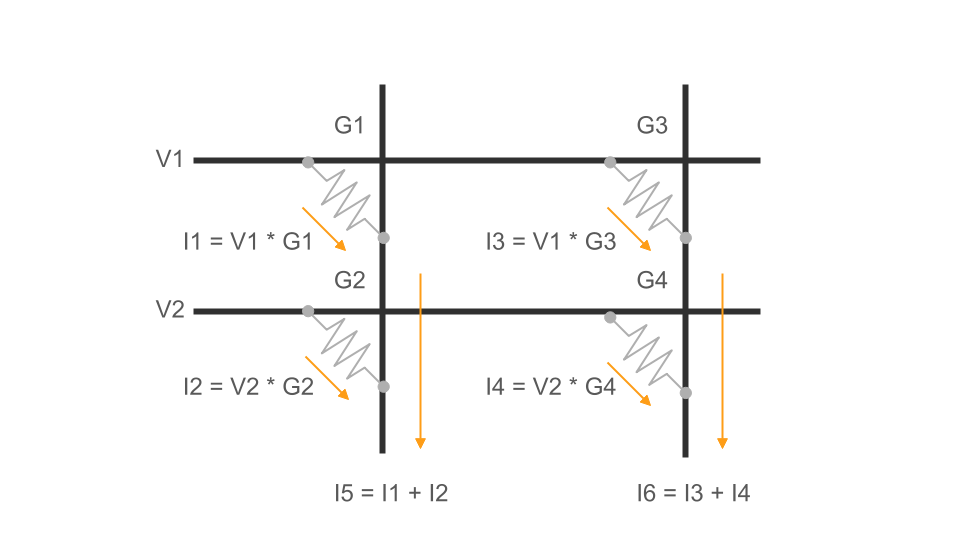}
    \caption{Representation of MAC computation a ReRAM crossbar array}
    \label{fig:reram}
\end{figure}

\section{Reliability Issue in ReRam and SNN}

While there have been some studies on ReRAM-based accelerators integrated with SNNs \cite{8993431} \cite{10118061} \cite{10.1145/3061639.3062311}, many of these works tend to overlook the reliability issues that both ReRAM and SNN devices face in real-world scenarios. These reliability issues can significantly impact the accuracy and performance of ReRAM with SNN devices when deployed in practical applications. This section discusses the unreliable factors associated with ReRAM and SNN devices and summarizes potential solutions proposed in previous research.

\subsection{Reliability Issue in ReRam}

\subsubsection{Device-Level Variation}:

In ReRAM technology, the hardware parameters exhibit a wide range of statistical distributions \cite{Hsu2015ASO} \cite{C8FD00106E} \cite{8360157} \cite{7495087} \cite{7838348} due to the stochastic nature of cell characteristics. This variability is primarily caused by changes in the shape of the conductive filament each time the cell is reprogrammed, leading to alterations in the ReRAM cell's LRS and HRS resistances. These variations in parameters can significantly affect the reliability and accuracy of devices that rely on ReRAM. The resistance variation in ReRAM cells makes it challenging to guarantee a one-to-one mapping of a single digital value to a fixed analog value. Additionally, it is difficult to ensure that a predefined resistance state can be precisely programmed into a ReRAM cell. Device measurement results have shown that the resistance distribution of ReRAM follows a normal distribution or a log-normal distribution\cite{8587661} \cite{7092614}. Due to these characteristics, ReRAM cells in LRS and HRS states exhibit a wide range of distributed resistance states, as depicted in Figure 6. Assuming an input bit is '1' (digital), this input value must first be converted into an analog voltage applied to the ReRAM cell. This analog value induces the ReRAM cell to produce a current output. However, since LRS and HRS states have a wide range of resistance values, the output current will also show a distribution, as illustrated in Figure 6(d).

\begin{figure}[h]
    \centering
    \includegraphics[width=1\textwidth]{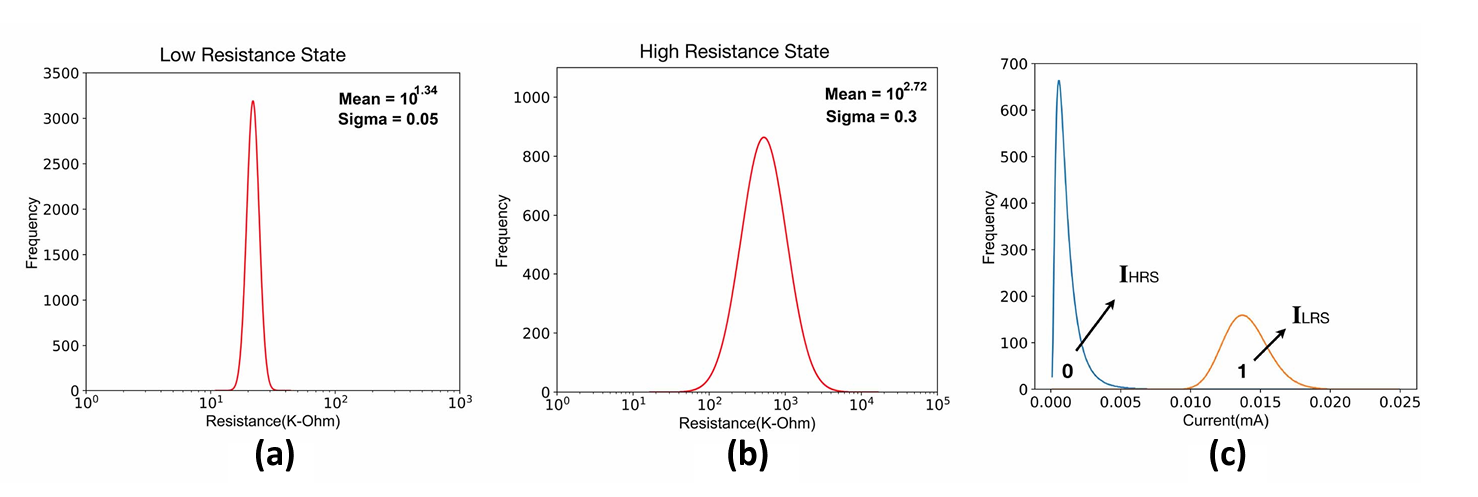}
    \caption{(a) Distribution of LRS by sampling ReRAM cells. (b) Distribution of HRS by sampling ReRAM cells. (c) Distribution of $I_{LRS}$ and $I_{HRS}$ for a ReRAM cell. (figure adapted from \cite{9211412})}
    \label{fig:variation}
\end{figure}

The distribution of output currents in ReRam-based crossbar arrays can lead to a significant challenge during the analog-to-digital conversion phase of MAC operations. Due to this variability, outputs that should be distinct may end up being interpreted as identical, resulting in multiple digital values being mapped to the same current value. This issue is known as overlapping error.\cite{9211412}

Furthermore, when a ReRAM cell is in its off state, ideally, no matter what the input voltage is—whether "0" or some other value—the output current should be zero. However, in practical applications, the resistance of a ReRAM cell in the off state rarely resets to infinity. As a result, as long as the input voltage is not zero, an off state ReRAM cell will still produce some output current \cite{8641288}. As depicted in Figure 6(d), the current in the High Resistance State ($I_{HRS}$) is not zero. This non-ideal behavior becomes especially problematic as the on-off ratio ($I_{on}/I_{off}$) decreases. A smaller on-off ratio means the difference between the current in the on state and the current in the off state is reduced, thus, the impact of the off-state current on the overall output current becomes more significant, leading to more severe overlapping errors.

When a ReRAM device has more activated word lines, the severity of overlapping errors increases significantly. This is due to the cumulative effect of multiple currents from activated word lines combining in a manner that creates more overlapping states between different intended digital outputs. As depicted in Figure 7(a), when there are three activated word lines, the possible output currents can manifest in only four cases: $3I_{HRS}$, $2I_{HRS} + 1I_{LRS}$, $1I_{HRS} + 2I_{LRS}$, and $3I_{LRS}$. These will convert to four outputs (0, 1, 2, 3). However, both $1I_{HRS} + 2I_{LRS}$ and $3I_{LRS}$ may output the same current, making it impossible to accurately discern whether the digital value should be 2 or 3 when the output current is identical.

As the number of activated word lines increases, the probability of overlapping errors rises. This is shown in Figures 7(b) and 7(c), which illustrate the scenarios with 10 and 20 activated word lines, respectively. The potential for overlapping errors in Figure 7(c) is much higher than in Figure 7(a), with a much larger area of current state overlaps. Additionally, the more activated word lines there are, the greater the deviation in misjudgment becomes. For example, as seen in Figure 7(c), the possible interpretations for $10I_{HRS} + 10I_{LRS}$ could span five different values (8, 9, 10, 11, 12).

\begin{figure}[h]
    \centering
    \includegraphics[width=1\textwidth]{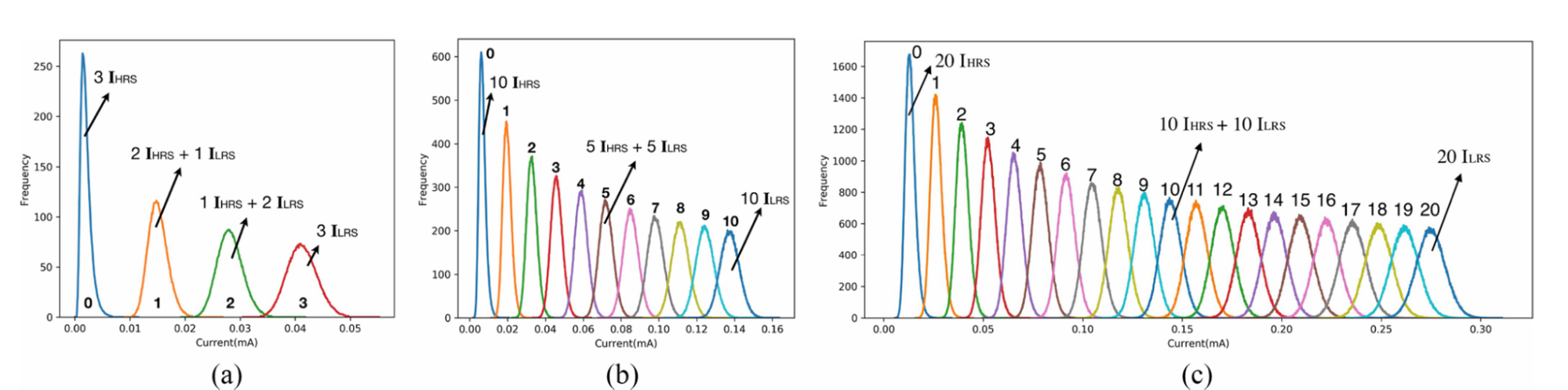}
    \caption{Reason and severity of the overlapping variation error. (a) Accumulated current distribution of three valid wordlines. (b) Accumulated current distribution of ten valid wordlines. (c) Accumulated current distribution of 20 valid wordlines. (figure adapted from \cite{9211412})}
    \label{fig:overlapping}
\end{figure}

Currently, numerous studies have proposed methods for improving overlapping analog variation errors,In the paper \cite{9211412} the authors propose three specific methods designed to alleviate the impact of overlapping analog variation errors. These methods are termed Weight Rounding Design(WRD), Adaptive Input Subcycling Design(AISD), Bitline Redundant Design(BRD).

\subsubsection{Weight Rounding Design}:

From Figure 7, we observe that the more LRS states a ReRAM cell has, the higher the probability of overlapping errors occurring. This is because, according to Ohm's law, the current value is inversely proportional to the resistance value under the same input voltage. Therefore, as shown in Figure 6, when a ReRAM cell is in the LRS state, the distribution of its output current is wider. This means that the more LRS states there are in the ReRAM, the wider the distribution of output currents will be. A wider distribution of output currents increases the probability of overlap with adjacent distributions, thereby leading to more severe overlapping errors.

In neural networks, model weights are first decomposed into binary form before being programmed into ReRAM cells, with the "1" in ReRAM representing the LRS state. Thus, the fewer "1" bits there are, the fewer LRS states are present in the ReRAM cell. To address this, the authors propose a method called Weight Rounding Design (WRD), which converts the original weight value to the nearest value with fewer "1" bits. For example, if a weight value is 255, its binary form is 0b011111111. Using WRD, 255 would be converted to 256, which has a binary form of 0b100000000. This method reduces the number of LRS states in ReRAM cells without significantly altering the weight value. This approach is based on the premise that DNNs are an approximate computing technique, meaning that changes in weight values within a certain range do not significantly decrease the accuracy of the DNN.

\subsubsection{Adaptive Input Subcycling Design}:

Due to the increased likelihood of overlapping variation errors when too many word lines are activated simultaneously, the authors proposed an input subcycling design aimed at reducing these errors by dividing an input cycle into multiple subcycles. When an input cycle requires n word lines to be completed, dividing it into m subcycles allows each subcycle to have only $\frac{n}{m}$ word lines activated. This approach reduces the probability of overlapping variation errors at the cost of increased total execution time.	

To implement this concept, they first introduced a Static Input Subcycling Design (SISD), which examines all potential values of m in an offline manner and determines a suitable fixed m for all input cycles. This method allows for a balanced trade-off between accuracy and performance by finding the appropriate n and m values. 

However, SISD does not account for the varying importance of the input cycles entering the ReRAM. Since the digital value of the input must first be converted into binary form, with bits “0” or “1” placed into the wordline, the impact of errors occurring at the Most Significant Bit (MSB) and Least Significant Bit (LSB) in the binary form can differ significantly. An error in the MSB during a MAC operation may result in being plus or minus 1, but when converted back to a digital value, the variation can be exponentially magnified.

Therefore, the authors proposed an Adaptive Input Subcycling Design (AISD), which adaptively splits each input cycle based on the relevant bit positions. In this scheme, the i-th cycle is divided into subcycles with $m_{i}$ wordlines. This adaptive approach strategically reduces the number of word lines activated during the MAC operations for the MSB to decrease the probability of overlapping errors. Conversely, it increases the number of word lines used during operations for the LSB to boost efficiency.By doing so, AISD can significantly enhance the accuracy and reliability of MAC operations.

\subsubsection{Bitline Redundant Design}:

The design concept behind Bitline Redundant Design (BRD) is to use additional storage space to reduce the likelihood of overlapping errors. BRD achieves this by employing redundant bit lines to perform calculations multiple times and then aggregating and averaging all computed results. This approach helps to minimize the impact of overlapping errors on the output current.

This effectiveness is based on the assumption that the distribution of output current follows a normal or log-normal distribution. By conducting multiple operations and averaging the results, the final output current is more likely to converge towards the central value of the distribution. Thus, the more redundant bit lines used, the more robust the output result is.

\subsubsection{Dynamical Fixed Point Data Representation}:

In \cite{8715178} the authors introduce a concept called Dynamical Fixed Point (DFP) data representation. DFP is a method of adjusting the decimal point of data based on the range of the data. The advantage of DFP is its ability to minimize the occurrence of zeros in the data representation without altering the actual value of the data. By dynamically shifting the decimal point to the left until there are no unused MSBs, DFP reduces the number of bit lines that remain inactive (or in the off state).

\subsubsection{Device-Variation-Aware Training}:

In \cite{8715178} the authors introduce a methodology known as Device-Variation-Aware training (DVA). The fundamental idea behind DVA is to preemptively inject weight variation during the training phase of DNNs. This strategy enables the resultant DNN model to be inherently cognizant of device variation, thereby enhancing its reliability in real-world applications involving variable hardware conditions. Additionally, a similar concept is explored in \cite{10191804} where the authors employ Error-Aware training. This approach involves the initial sampling of the variation distribution followed by the incorporation of these variations into the weight parameters during training. 
These strategies enable the resultant DNN model to be inherently cognizant of device variation, improving the robustness of neural networks and effectively handling real-world operational noise and discrepancies typical of non-volatile memory technologies.

However, the methods proposed in these two paper do not provide detailed descriptions of how they establish fault maps or conduct simulations, nor do they mention whether they simulate the impact of off-state currents. These factors can significantly influence experimental outcomes. Therefore, I offer some perspectives on how to conduct more comprehensive simulations.

Currently, most papers on error injection involve summing the values from bit lines and then modifying them according to a sampled distribution. A more rigorous approach would be to first convert weights into binary and adjust the current output from each cell according to the distribution function of $I_{LRS}$, followed by aggregation. However, these methods typically do not account for the impact of off-state currents. This is critical because cells with a binary weight bit of 0, where the resistance cannot be set to infinity, would still produce a current when the input voltage is not zero.

I believe that in simulations when the binary weight bit is 0, it should be adjusted according to the distribution of $R_{HRS}$. Similarly, for binary weight bits set to 1, adjustments should be made based on the distribution of $R_{LRS}$. Subsequently, these adjusted values would undergo multiplication and summation. This approach would more accurately simulate the effects of weight distribution and the impact of off-state currents, thus providing a more realistic portrayal of how device variability could affect the performance of DNNs in ReRAM-based systems. This enhanced simulation methodology could lead to better strategies for managing errors and improving the robustness and accuracy of neural networks in practical settings.

\subsection{Reliability Issue in SNN Hardware Accelerator}

An SNN hardware neuron cell can broadly be categorized into two main components: (1) Synapse and (2) Neuron. The synapse component of the hardware is used to store the connection strength, or weight, between two neurons. The neuron component typically employs an I\&F model, commonly referred to as the I\&F circuit. The I\&F circuit receives the output from the synapse and uses it to compute the membrane potential ($V_{mem}$), ultimately deciding whether or not to output a spike.

SNN hardware is susceptible to transient faults\cite{1438282} \cite{9159745}, often caused by high-energy particle strikes, which can disrupt hardware operations. In the neuron part of the hardware, four potential issues can arise: (1) $V_{mem}$ cannot increase; (2) $V_{mem}$ cannot leak (decrease); (3) $V_{mem}$ cannot reset; (4) the neuron is unable to produce a spike.

In the study \cite{10.1145/3489517.3530657} the authors introduce the Bound-and-Protect (BnP) techniques. The concept behind BnP involves using specialized hardware to monitor for faulty $V_{mem}$ reset conditions by observing the comparison output of $V_{mem} > V_{thr}$. If the output is true for an extended period, the neuron's spike generation is disabled to prevent the occurrence of burst spikes. The specialized hardware includes radiation-hardened registers and hardened combinational logic units, which are designed to resist high-energy particle strikes and prevent soft errors.

\section{The Combiantion of SNN and NVM}

In the publication \cite{9405141} the authors develop a VGG7 SNN model. The specific input encoding formula is:

\begin{center}
$X_{i} = \alpha V_{d}Z$, where $Z \sim Bernoulli(\frac{x_{i}}{x_{MAX}})$

$\alpha = \frac{x_{MAX}}{V_{d}}$
\end{center}

In this formula, $V_{d}$ represents the drain voltage, $V_{d}Z$ serves as a random variable input of a spiking neuron. Z is a random variable following Bernoulli distribution which is bi-state and the probability of 1 is proportional to the strength of input, $x_{i}$. The scaling factor between input neural activities and input voltage is $\alpha$

The duration from the initiation of input to the first spike generated by a neuron is transformed into a numerical value through a sum-of-products operation, thereby converting the timing of neural responses to quantifiable outputs.

Given the random nature of $V_{dZ}$, the input to the SNN can be regarded as a random variable, implying that the dynamics of the SNN are a stochastic process. This characterization allows for analyzing the relationships among computing precision, circuit parameters, and array current in a stochastic framework.

The authors conducted simulations by integrating the SNN model with various non-volatile memory types and analyzed the results based on the setup and outcomes of the simulations. They found that a smaller on-off ratio necessitates a larger membrane capacitor to achieve high accuracy. Conversely, when the on-off ratio exceeds 1000, performance closely approaches that of an ideal scenario.

High on-currents require larger membrane capacitors to maintain accuracy, as excessive on-current can lead to rapid saturation of the membrane capacitor and prematurely reach the threshold, resulting in earlier firing of the first spike, which can diminish output precision.

When the normalized standard deviation of the cell current exceeds 10\%, accuracy begins to degrade. This variation can be mitigated by employing weight duplication, which involves repeating calculations for the same weight, summing the results, and then averaging them. This method reduces the area of output current overlapping. However, weight duplication also increases the overall output current. Consequently, while weight duplication helps in reducing overlapping error, it requires a larger membrane capacitor and more computing energy to accommodate the increased current demands.

\section{Conclusion}

Artificial intelligence has been widely applied across various domains to date. Current AI models are predominantly deployed on large cloud-computing systems. However, for edge devices and Internet of Things (IoT) applications, there is a need for AI models that require less energy and reduced computing resources. Methods such as model compression and quantization have been explored, but ANN models still cannot achieve brain-like efficiency. In contrast, brain-inspired neuromorphic computing methods, such as SNNs, can achieve accuracy close to ANNs with much lower energy consumption. Yet, most SNNs that reach comparable accuracy to ANNs often rely on backpropagation through time or ANN to SNN conversion, which minimally utilizes the rich timing information inherent in spikes. Therefore, there is a need to develop new learning algorithms that can better harness this timing information to achieve higher efficiency with lower computing resources. Additionally, using Compute-in-Memory (CIM) architectures could potentially overcome the Von-Neumann bottleneck, enhancing efficiency and reducing energy consumption, making the integration of SNNs and non-volatile memory a likely future trend for edge device AI.

While SNNs with NVM appear promising in ideal environments, they face reliability issues in real-world applications. For example, in ReRAM hardware, the shape of each cell’s conductance filament may change with every reprogram, leading to imprecise resistance values. Moreover, setting a cell’s value to zero requires programming its resistance to infinity, which is unachievable in practice, hence leading to off current issues. The hardware components of SNNs' I\&F circuits may also experience transient faults due to high-energy particle strikes, leading to operational errors.

Therefore, researching how to resolve reliability issues in SNNs and ReRAM is crucial for their practical deployment. Several methods have been proposed to improve the reliability of SNNs and ReRAM, such as using radiation-hardened hardware to detect errors in I\&F circuits or employing weight duplication to mitigate weight variation. However, most current research on ReRAM hardware relies on simulations, and real-world environments might introduce new challenges.

Moreover, most studies discuss SNN and NVM hardware separately, whereas "Robust Brain-Inspired Computing: On the Reliability of Spiking Neural Network Using Emerging Non-Volatile Synapses" integrates and analyzes SNNs with different NVMs, deriving relationships between the first time to fire, output precision, membrane capacitor, and on and off currents. However, this paper's derivations are based on specific encoding and decoding methods used in SNNs, and since there are many ways to encode and decode in SNNs, a more general analysis of combinations is desired.

\bibliographystyle{ACM-Reference-Format}
\bibliography{survey}

\end{CJK}
\end{document}